\documentclass[10pt,twocolumn,letterpaper]{article}

\usepackage[pagenumbers]{cvpr} 

\usepackage[dvipsnames]{xcolor}

\newcommand{\norm}[1]{\left\lVert#1\right\rVert}

\usepackage{makecell}

\definecolor{cvprblue}{rgb}{0.21,0.49,0.74}
\usepackage[pagebackref,breaklinks,colorlinks,citecolor=cvprblue]{hyperref}

\def\paperID{408} 
\def\confName{3DV\xspace}
\def\confYear{2025\xspace}
\newcommand{\method}{Splatt3R\xspace}

\newcommand{\bu}{\boldsymbol{u}}
\newcommand{\bmu}{\boldsymbol{\mu}}

\makeatletter
\renewcommand{\paragraph}{%
  \@startsection{paragraph}{4}%
  {\z@}{-0.5em}{-0.5em}%
  {\normalfont\normalsize\bfseries}%
}
\makeatother

\title{\method: Zero-shot Gaussian Splatting from Uncalibrated Image Pairs}

\author{
    Brandon Smart\textsuperscript{1} \qquad
    Chuanxia Zheng\textsuperscript{2} \qquad
    Iro Laina\textsuperscript{2} \qquad
    Victor Adrian Prisacariu\textsuperscript{1} \\
    \textsuperscript{1}Active Vision Lab, University of Oxford \qquad
    \textsuperscript{2}Visual Geometry Group, University of Oxford\\
    {\tt\small \{brandon, cxzheng, iro, victor\}@robots.ox.ac.uk}
}

\begin{document}

\twocolumn[{%
\renewcommand\twocolumn[1][]{#1}%
\maketitle
\begin{center}
    \captionsetup{type=figure}
    \vspace{-0.1cm}
    \includegraphics[width=0.9\textwidth]{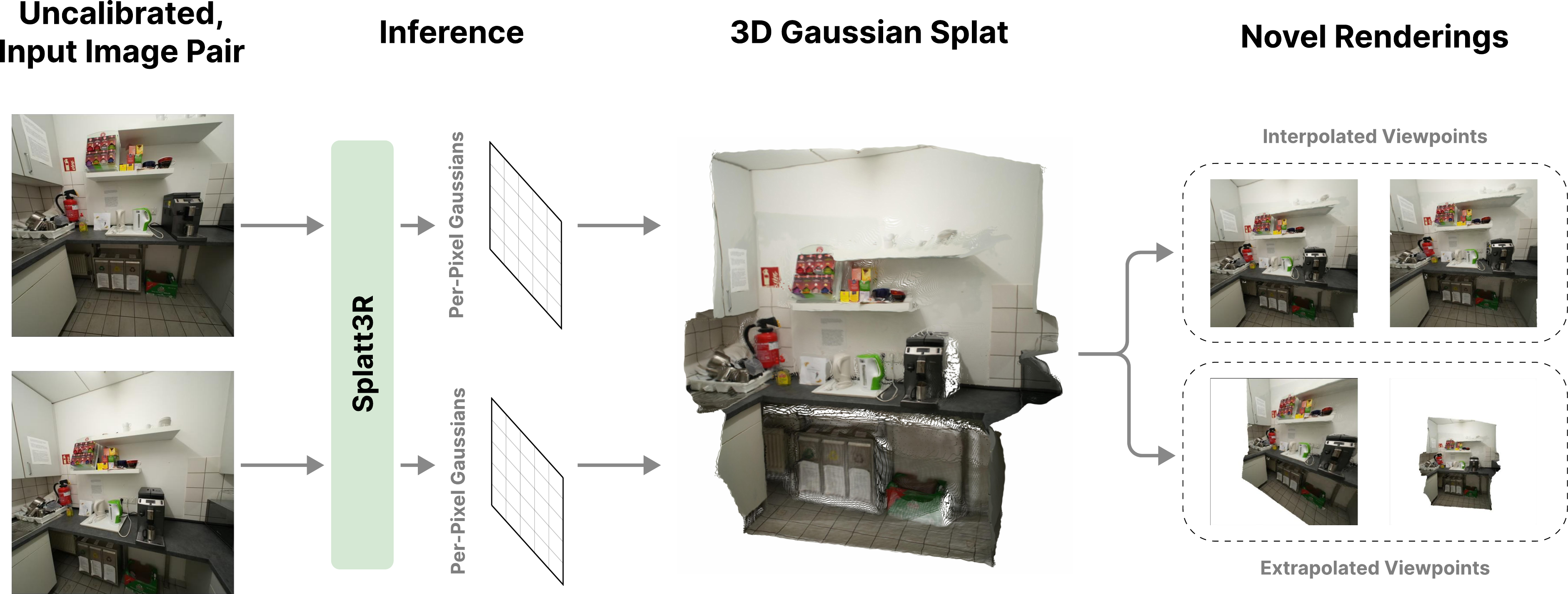  }
    \captionof{figure}{We introduce \textit{\method}, a feed-forward model that can directly predict a 3D Gaussian Splat from a stereo pair of images with unknown camera parameters. 
    We base our work on MASt3R, and following their simple architecture, we avoid any explicit prediction of camera poses, intrinsics or monocular depth.
    \method can perform both interpolation and extrapolation of novel views from the input.
    }
    \label{fig:overview}
    \vspace{0.6cm}
\end{center}
}]

\maketitle

\begin{abstract}
\vspace{-.3cm}
In this paper, we introduce \method, a pose-free, feed-forward method for in-the-wild 3D reconstruction and novel view synthesis from stereo pairs.
Given uncalibrated natural images, \method can predict 3D Gaussian Splats without requiring any camera parameters or depth information.
For generalizability, we build \method upon a ``foundation'' 3D geometry reconstruction method, MASt3R,
by extending it to deal with both 3D structure and appearance.
Specifically, unlike the original MASt3R which reconstructs only 3D point clouds, we predict the additional Gaussian attributes required to construct a Gaussian primitive for each point.
Hence, unlike other novel view synthesis methods, \method is first trained by optimizing the 3D point cloud's geometry loss, and then a novel view synthesis objective.
By doing this, we avoid the local minima present in training 3D Gaussian Splats from stereo views.
We also propose a novel loss masking strategy that we empirically find is critical for strong performance on extrapolated viewpoints.
We train \method on the ScanNet++ dataset and demonstrate excellent generalisation to uncalibrated, in-the-wild images.
\method can reconstruct scenes at 4FPS at $512 \times 512$ resolution, and the resultant splats can be rendered in real-time.
\end{abstract}
\section{Introduction}

We consider the problem of 3D scene reconstruction and novel view synthesis from sparse, \emph{uncalibrated natural images} in just one forward pass of a trained model.
While recent breakthroughs have been made in 3D reconstruction and novel view synthesis by using neural scene representations, \eg. SRN~\cite{sitzmann2019scene}, NeRF~\cite{mildenhall2020nerf}, LFN~\cite{sitzmann2021light}, and non-neural scene representations, \eg. 3D Gaussian Splatting (3D-GS)~\cite{kerbl2023gaussian}, these methods are far from being accessible to casual users, due to expensive, iterative, per-scene optimization procedures, which are often slow and are unable to utilize learned priors from training datasets.
More importantly, reconstruction quality is poor when trained from only a pair of stereo images, as these methods require a dense collection of dozens or hundreds of images to produce high-quality results.

To mitigate these issues, generalizable 3D reconstructors~\cite{yu2021pixelnerf,wang2021ibrnet,johari2022geonerf,chibane2021stereo,chen2021mvsnerf,du2023cross}, aim to predict pixel-aligned features for radiance fields from sparse \emph{calibrated images} using feed-forward networks.
These models are trained by differentiably rendering the predicted, parameterized representations from target viewpoints and supervising them with ground truth images captured from the same camera pose.
By learning priors across large datasets of input scenes, these models avoid the failure cases of traditional per-scene optimization from sparse images.
To avoid the expensive volumetric processing in NeRF, several feed-forward Gaussian Splatting models~\cite{charatan2024pixelsplat,zheng2024gpsgaussian,chen2024mvsplat,szymanowicz2024splatter,szymanowicz2024flash3d} have been proposed to explore 3D reconstruction from sparse views.
They use a cloud of pixel-aligned 3D Gaussian primitives~\cite{kerbl2023gaussian} to represent the scene.
The 3D locations of these Gaussian primitives are parameterized using their depth along the ray, which is explicitly calculated using the \emph{known intrinsic and extrinsic camera parameters} from the input images.

Due to their reliance on known camera parameters, these methods can \emph{not} be directly used on ``in-the-wild'' uncalibrated images.
Ground truth poses are assumed to be available, or camera pose estimation is implied as a pre-processing step\,---\, existing methods are typically tested on datasets where poses have been reconstructed by running SfM software on dozens or hundreds of images of the same scene. Methods which attempt to use SfM or multi-view stereo (MVS) pipelines typically use a string of algorithms for matching points, triangulating them, finding essential matrices, and estimating camera extrinsics and intrinsics.

In this paper, we introduce \method, a feed-forward model that takes as input two uncalibrated images, and outputs 3D Gaussians to represent the scene. Specifically, we use a feed-forward model to predict pixel-aligned 3D Gaussian primitives for each image, and then render novel views using a differentiable renderer. We achieve this without relying on any additional information such as camera intrinsics, extrinsics, or depth.

Without explicit pose information, one key challenge is identifying where to place the 3D Gaussian centers. Even with pose information, iterative 3D Gaussian Splatting optimization is susceptible to local minima~\cite{kerbl2023gaussian, charatan2024pixelsplat}. Our solution is to jointly address the lack of camera poses and the problem of local minima by explicitly supervising and regressing the ground truth 3D point clouds for each training sample. In particular, we observe that the architecture used to produce MASt3R's pixel-aligned 3D point clouds ~\cite{leroy2024grounding} closely aligns with the existing pixel-aligned 3D Gaussian splatting architectures using in feed-forward Gaussian methods~\cite{charatan2024pixelsplat,chen2024mvsplat,szymanowicz2024splatter,szymanowicz2024flash3d}. Therefore, we seek to show that simply adding a Gaussian decoder to a large-scale pre-trained ``foundation'' 3D MASt3R model, without any bells and whistles, is sufficient to develop a \emph{pose-free}, \emph{generalizable} novel view synthesis model.

One notable limitation of most existing generalizable 3D-GS methods is that they only supervise novel viewpoints which are between the input stereo views~\cite{charatan2024pixelsplat,chen2024mvsplat}, rather than learning to extrapolate to farther viewpoints. The challenge with these extrapolated viewpoints is that they often see points that are obscured to the input camera views, or are outside of their frustums entirely. Thus, supervising the novel view rendering loss for these points is counterproductive, and can be destructive to the model's performance. By only supervising the novel view rendering loss for views that are between the two context images, existing works avoid attempting to reconstruct many unseen parts of the scene. However, this means that the model is not trained to accurately generate novel view renderings for views beyond the stereo baseline. To address this, we employ a loss masking strategy based on frustum culling and covisibility testing, calculated using the ground truth poses and depth maps known during training. We apply mean squared error and LPIPS loss only to the parts of the rendering that can be feasibly reconstructed, preventing updates to our model from unseen parts of the scene. This allows training with wider baselines, and for supervising novel views that are beyond the stereo baseline.

We present, for the first time, a method that predicts 3D Gaussian Splats for scene reconstruction and novel view synthesis from a pair of unposed images in a single forward pass of a network. We construct baselines out of existing work and show that our method surpasses them in visual quality and perceptual similarity to the ground truth images. More impressively, our trained model is capable of generating photorealistic novel view synthesis from in-the-wild uncalibrated images. This significantly relaxes the need for dense image inputs with precise camera poses, addressing a major challenge in the field.
\section{Related Work}

\subsection{Novel View Synthesis}

Many representations have been used for 3D Novel View Synthesis (NVS),
such as luminagraphs~\cite{gortler1996lumigraph}, light fields~\cite{levoy1996light}, and plenoptic functions~\cite{adelson1991plenoptic}. Neural Radiance Fields (NeRFs) have achieved photo-realistic representations of 3D scenes using view-dependent, ray-traced radiance fields, encoded by neural networks trained through per-scene optimization on densely collected image sets~\cite{mildenhall2020nerf, muller2022instant, barron2022mipnerf360}. Recently, 3D Gaussian Splatting~\cite{kerbl2023gaussian} has greatly increased the training and rendering speed of radiance fields by training a set of 3D Gaussian primitives to represent the radiance of each point in space, and rendering them through an efficient rasterization process.

To avoid intensive per-scene optimization, generalizable NVS pipelines have been developed, which infer 3D representations directly from multi-view images~\cite{wang2021ibrnet, yu2021pixelnerf, reizenstein2021common, chibane2021stereo, chen2021mvsnerf, suhail2022generalizable, johari2022geonerf, liu2022neural, smith2023flowcam, zheng2024gpsgaussian, li2024ggrt, liu2024fast, wewer2024latentsplat}. Rather than performing per-scene optimization, these methods are trained across large datasets of scenes, allowing data-driven priors to be learned that can ground reconstruction for newly observed scenes. By leveraging these data-driven priors, these methods have evolved to work with sparse image sets~\cite{long2022sparseneus, ni2024colnerf, li2024taming, chen2024mvsplat} and even stereo image pairs~\cite{zhou2018stereo, du2023cross, charatan2024pixelsplat, lee2024generalizable}, significantly reducing the number of reference images required to obtain a radiance field for NVS.

Recent methods, such as pixelSplat~\cite{charatan2024pixelsplat}, MVSplat~\cite{chen2024mvsplat}, GPS-Gaussian~\cite{zheng2024gpsgaussian}, SplatterImage~\cite{szymanowicz2024splatter} and Flash3D~\cite{szymanowicz2024splatter} use a cloud of 3D Gaussian primitives placed along camera rays explicitly calculated from camera parameters, aiming to predict one (or multiple) 3D Gaussian primitives per-pixel in each image. However, these existing methods assume the availability of camera intrinsics and extrinsics for each image at testing time, which limits their applicability to in-the-wild photo pairs. Many methods have been proposed for per-scene optimization with unknown camera poses~\cite{wang2021nerfmm, lin2021barf, jeong2021self, bian2023nope, bian2023porf}, however these depend on large collections of images. Recent studies propose methods to jointly predict camera parameters and 3D representations in a generalizable manner, although these are limited to sparse setups ~\cite{truong2023sparf, jiang2023leap}. In contrast, we propose \method to address the gap in generalizable stereo NVS with unknown camera parameters. Among closely related works, FlowCam~\cite{smith2023flowcam} removes the need for pre-computed cameras using dense correspondences from optical flow, but it requires sequential input and shows limited rendering performance.

By integrating the recent stereo reconstruction work MASt3R with 3D Gaussians, our method effectively handles larger baselines without the need for pre-processed cameras. GGRt~\cite{li2024ggrt} also seeks to model 3D Gaussian Splats without known camera poses or intrinsics, but instead focuses on processing video sequences with small baselines between frames, introducing caching and deferred back-propogation techniques to aid reconstruction from long video sequences. DBARF~\cite{chen2023dbarf} also aims to jointly learn camera poses and reconstruct radiance fields, but uses a NeRF-based approach and focuses on calculating poses using cost maps derived from learned features.

\subsection{Stereo Reconstruction}

Traditionally, the stereo reconstruction task involves a sequence of steps. Starting with keypoint detection and feature matching~\cite{harris1988combined, trajkovic1998fast, lowe2004distinctive, detone2018superpoint}, camera parameters are estimated using fundamental matrices~\cite{zhang1995robust, luong1996fundamental, ranftl2018deep}.  Next, dense correspondence is established through epipolar line search~\cite{barnard1982computational, kanade1996stereo, ishikawa1998occlusions} or stereo matching~\cite{zbontar2015computing, chang2018pyramid, zhang2019ga}, enabling the triangulation of 3D points~\cite{hartley2004sub, hartley1992stereo, hartley1997triangulation}. This process can be optionally refined by photometric bundle adjustment~\cite{delaunoy2014photometric,woodford2020large}. With the advent of deep learning, numerous methods have been proposed to integrate certain steps, such as joint depth and camera pose estimation, and optical flow~\cite{garg2016unsupervised, godard2017unsupervised, zhan2018unsupervised, yin2018geonet, chi2021feature, wang2021deep}. However, all these methods rely on explicit correspondence, making them challenging to apply when the overlap between images is limited.

Recently, DUSt3R~\cite{wang2024dust3r} introduced an innovative approach to address this challenge by predicting point maps for a pair of uncalibrated stereo images in one coordinate system with implicit correspondence searching. The follow-up paper MASt3R~\cite{leroy2024grounding} primarily focuses on improvements to image matching, but improves on DUSt3R by predicting points in metric space and achieving greater accuracy. These methods have shown promising stereo reconstruction results even when there is little or no overlap between the images. While the raw point maps are sufficiently accurate for several downstream applications like pose estimation,
they are not designed to be directly rendered. In contrast, our method augments MASt3R to predict 3D Gaussian primitives, which enables fast and photo-realistic NVS.
\section{Method}
\begin{figure*}
  \centering
  \includegraphics[width=1.0\textwidth]{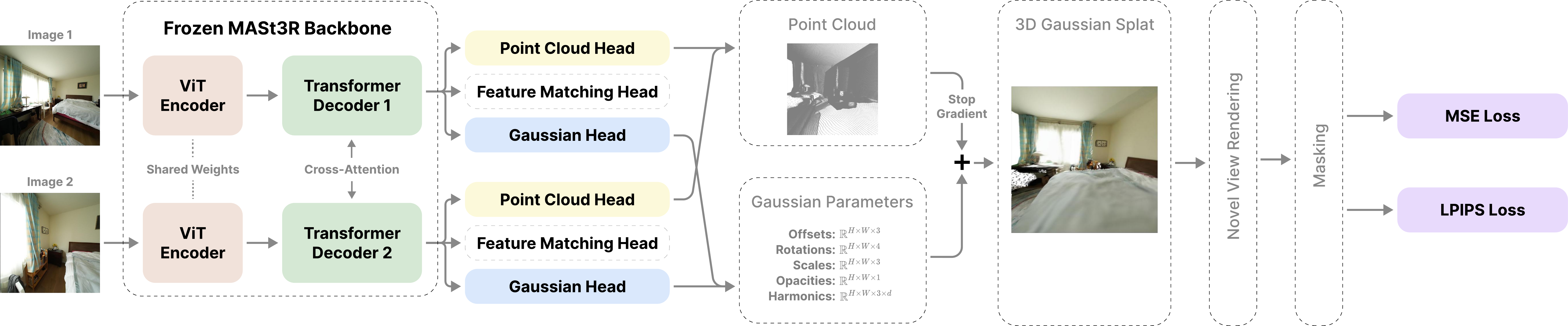}
  \caption{\textbf{Method overview.}
  We encode the two uncalibrated images using MASt3R's pretrained ViT encoder and cross-attention transformers, which we freeze during training. In addition to MASt3R's prediction head for point positions and confidences, we introduce a Gaussian head that predicts offsets, spherical harmonics, rotations, scales and opacities. We supervise novel renderings of the generated 3D Gaussian Splats using mean squared error (MSE) and LPIPS.
  } 
  \label{fig:method}
\end{figure*}

Given two uncalibrated images $\mathcal{I}=\{{\mathbf{I}}^{i}\}_{i=\{1, 2\}}$, (${\mathbf{I}}^i \in \mathbb{R}^{H \times W \times 3}$), our goal is to learn a mapping $\Phi$ that takes as input
$\mathcal{I}$ and outputs 3D Gaussian parameters for both geometry and appearance. 
We achieve this by simply adding a third branch to MASt3R to output the additional attributes required for 3D Gaussians.
Before outlining the details of our proposed method,
we provide a brief overview of 3D Gaussian Splatting in \cref{sec:gaussian_splatting}, followed by an overview of MASt3R in \cref{sec:mast3r}. We then describe how we modify the MASt3R architecture to predict 3D Gaussian Splats for novel view synthesis in \cref{sec:gaussian_prediction}. Finally, we outline our training and evaluation protocols in \cref{sec:protocols}.

\subsection{3D Gaussian Splatting}
\label{sec:gaussian_splatting}

\paragraph{Scenes as sets of 3D Gaussians.}
We begin by briefly reviewing 3D Gaussian Splatting (3D-GS)~\cite{kerbl2023gaussian}.
3D-GS represents the radiance field of a scene using a set of anisotropic, 3D Gaussians, each of which represents the radiance emitted in the spatial region around a point. Each Gaussian is parameterized using its mean position $\bmu \in \mathbb{R}^3$, opacity $\alpha \in \mathbb{R}$, covariance $\Sigma \in \mathbb{R}^{3 \times 3}$ and view-dependent color $S \in \mathbb{R}^{3 \times d}$ (here parameterized using $d$-degree spherical harmonics). Like other works, we reparameterize the covariance matrix with a rotation quaternion $q \in \mathbb{R}^{4}$ and scale $s \in \mathbb{R}^{3}$ to ensure positive semi-definite covariance matrices.
In our experiments, we focus on constant, view-independent color for each gaussian ($S \in \mathbb{R}^3$), and ablate view-dependent spherical harmonics.
Original 3D-GS uses an iterative process to fit the Gaussian Splats to a single scene, but Gaussian primitives have vanishingly small gradients if the distance to their `correct' location is greater than a few standard deviations, and can often get stuck in local optima during optimization~\cite{charatan2024pixelsplat}. 3D-GS partially overcomes these problems using initialization from SfM point clouds and non-differentiable `adaptive density control' to split and prune gaussians~\cite{kerbl2023gaussian}. This method is effective, but requires a dense collection of images, and cannot be used for generalizable, feed-forward models, which directly predict Gaussians without per-scene optimization.

\paragraph{Feed-forward 3D Gaussians.}
Very recently, given a set of $N$ images $\mathcal{I}=\{{\mathbf{I}}^{i}\}_{i=1}^N$, the generalizable 3D-GS methods~\cite{charatan2024pixelsplat,chen2024mvsplat,szymanowicz2024splatter,szymanowicz2024flash3d} predict pixel-aligned 3D Gaussian Primitives. In particular, for each pixel $\bu=(u_x,u_y,1)$, the parameterized Gaussian is predicted with its opacity $\alpha$, depth $d$, offsets $\Delta$, covariance $\Sigma$ expressed as rotation and scale, and the parameters of the colour model $S$. The location of each Gaussian is given by $\bmu = K^{-1} \bu d + \Delta$, where $K$ is the camera intrinsics. Of particular note, pixelSplat predicts a probabilistic distribution over depth, which seeks to avoid the problem of local minima by tying the probabilistic density to the opacity of the Gaussian primitives sampled~\cite{charatan2024pixelsplat}. However, these parameterizations cannot be directly applied to 3D-GS prediction from \emph{uncalibrated images}, which have unknown camera rays. Instead, we directly supervise the positions of per-pixel Gaussian primitives using `ground-truth' point clouds. This allows the Gaussian corresponding to each pixel to have a direct path of monotonically decreasing loss leading to its correct position during training.

\subsection{MASt3R Training}
\label{sec:mast3r}

As discussed, we wish to directly supervise the 3D location of each pixel in a pair of uncalibrated images.
This task has recently been explored by DUSt3R~\cite{wang2024dust3r} (and its follow-up work MASt3R~\cite{leroy2024grounding}), a multi-view stereo reconstruction method that directly regresses a model for predicting 3D point clouds.
For simplicity, we collectively refer to these methods as `MASt3R' for the remainder of the paper.

Given two images $\textbf{I}^1, \textbf{I}^2 \in \mathbb{R}^{W \times H \times 3}$, MASt3R learns to predict the 3D locations for each pixel $\hat{X}^1, \hat{X}^2 \in \mathbb{R}^{W \times H \times 3}$, alongside corresponding confidence maps $C^1, C^2 \in \mathbb{R}^{W \times H}$. Here, the model aims to predict both point maps in the coordinate frame of the first image, which removes the need for transforming point clouds from one image's coordinate frame to the other using camera poses. This representation, like generalizable 3D reconstruction approaches, assumes the existence of a single, unique location where the ray corresponding to each pixel intersects with the surface geometry, and does not attempt to model non-opaque structures like glass or fog.

Given ground truth pointmaps $X^1$ and $X^2$, for each valid pixel $i$ in each view $v \in \{1, 2\}$, the training objective $L_{pts}$ is defined as:
\begin{equation}
    \label{dust3r_loss}
    \begin{split}
        L_{pts} = \sum_{v \in \{1, 2\}} \sum_{i} C_{i}^{v} L_{regr}(v, i) - \gamma \log(C_{i}^{v})
    \end{split}
\end{equation}
\begin{equation}
    \label{eqn:regr_loss}
    \begin{split}
        L_{regr}(v, i) = \norm{\frac{1}{z} X_{i}^{v} - \frac{1}{\bar{z}} \hat{X}_{i}^{v}}
    \end{split}
\end{equation}
$L_{pts}$ is a confidence-based loss used to handle points with ill-defined depths, such as points corresponding to the sky, or to translucent objects. The hyperparameter $\gamma$ governs how confident the network should be, while $z$ and $\bar{z}$ are normalization factors used for non-metric datasets (set to $z = \bar{z} = 1$ for metric datasets). In our experiments, we use a frozen MAST3R model pre-trained with this objective, and only apply novel view rendering losses during training. We experiment with fine-tuning using this loss in \cref{tab:ablation_table}.

\subsection{Adapting MASt3R for Novel View Synthesis}
\label{sec:gaussian_prediction}

We now present \textit{\method}, a feed-forward model that predicts 3D Gaussians from uncalibrated image pairs. 
Our key motivation derives from the conceptual similarity between MASt3R and generalizable 3D-GS models, such as pixelSplat~\cite{charatan2024pixelsplat} and MVSplat~\cite{chen2024mvsplat}.
First, these methods all use feed-forward, cross-attention network architectures to extract information between input views.
Second, MASt3R predicts pixel-aligned 3D points (and confidence) for each image, whereas generalizable 3D-GS models~\cite{charatan2024pixelsplat,chen2024mvsplat,szymanowicz2024splatter,szymanowicz2024flash3d} predict pixel-aligned 3D Gaussians for each image.
Thus,
we follow the spirit of MASt3R, and show that a simple modification to the architecture, alongside a well-chosen training loss, is sufficient to achieve strong novel view synthesis results.

Formally, given a set of uncalibrated images $\mathcal{I}$, MASt3R encodes each image $\mathcal{I}^{i}$ simultaneously using a vision transformer (ViT) encoder~\cite{dosovitskiy2020image}, then passes them to a transformer decoder which performs cross-attention between each image.
Normally, MASt3R has two prediction heads, one that predicts a 3D point ($x$) and confidence ($c$) for each pixel, and a second which is used for feature matching, which is not relevant to our task, and can be ignored. We introduce a third head, which we refer to as the \emph{`Gaussian head'}, that runs \emph{in parallel} to the existing two heads. This head predicts covariances (parameterized by rotation quaternions $q \in \mathbb{R}^4$ and scales $s \in \mathbb{R}^3$), spherical harmonics ($S \in \mathbb{R}^{3 \times d}$) and opacities ($\alpha \in \mathbb{R}$) for each point. Additionally, we predict an offset ($\Delta \in \mathbb{R}^3$) for each point, and parameterize the mean of the Gaussian primitive as $\mu = x + \Delta$. This allows us to construct a complete Gaussian primitive for each pixel, which we can then render for novel view synthesis. 

During training, we only train the Gaussian prediction head, relying on a pre-trained MASt3R model for the other parameters. Following MASt3R's point prediction head, we use the DPT architecture ~\cite{ranftl2021vision} for our Gaussian head. An overview of the model architecture is shown in \cref{fig:method}.

Following existing generalizable 3D-GS works, we use different activation functions for each Gaussian parameter type, including normalization for quaternions, exponential activations for scales and offsets, and sigmoid activations for opacities. Additionally, to aid in the learning of high-frequency color, we seek to predict the residual between each pixel's color and the color we apply to that pixel's corresponding Gaussian primitive. 

Following MAST3R's practice of predicting the 3D locations of all points in the first image's camera frame, the predicted covariances and spherical harmonics are considered as being in the first image's camera frame. This avoids the need to use ground truth transformations to convert these parameters between reference frames, which existing methods do~\cite{charatan2024pixelsplat}. The final set of Gaussian primitives is the union of the Gaussian primitives predicted from both images.

\subsection{Training Procedure and Loss Calculation}
\label{sec:protocols}

To optimize our Gaussian parameter predictions we supervise novel view renderings of the predicted scene, as in existing work~\cite{charatan2024pixelsplat,chen2024mvsplat,szymanowicz2024splatter}.
During training, each sample consists of two input `context' images which we use to reconstruct the scene, and a number of posed `target' images which we use to calculate rendering loss.

\begin{figure}
  \centering
  \includegraphics[width=0.9\columnwidth]{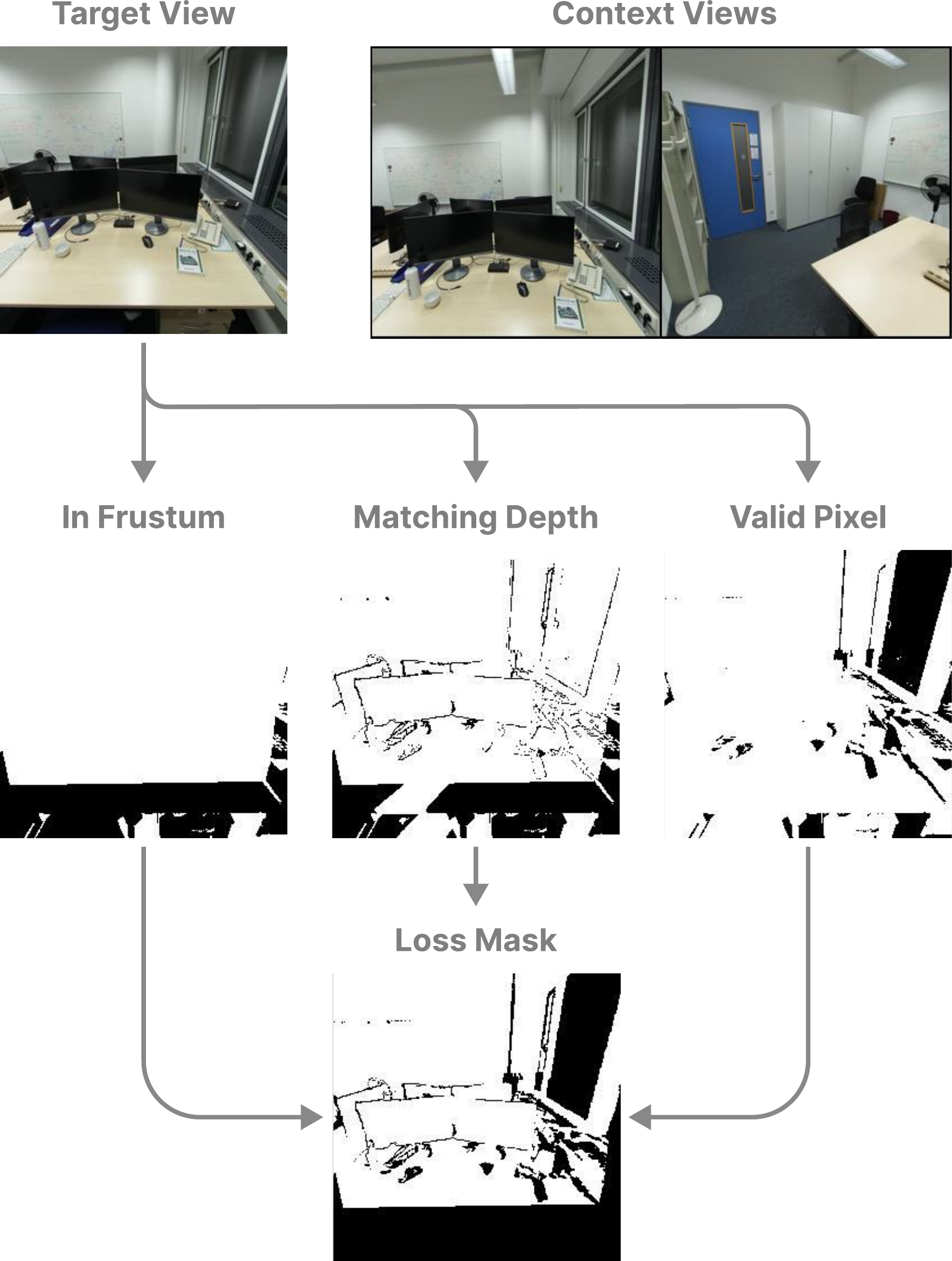}
  \caption{\textbf{Our loss masking approach.} Valid pixels are considered to be those that are: inside the frustum of at least one of the views, have their reprojected depth match the ground truth, and are considered valid pixels with valid depth in their dataset.} 
  \label{fig:loss_masks}
\end{figure} 

Some of these target images may contain regions of the scene that were not visible to the two context views due to being obscured, or outside of the context view frustums entirely. Supervising the rendering loss for these pixels would be counterproductive and potentially destructive to the model's performance. Existing generalizable, feed-forward radiance field prediction methods attempt to avoid this problem by only synthesizing novel views for viewpoints that are between the input stereo views~\cite{charatan2024pixelsplat, du2023cross, chen2024mvsplat}, reducing the number of unseen points that need to be reconstructed. Instead, we seek to train ours to extrapolate to farther viewpoints that are not necessarily an interpolation between the two input images. 

\begin{table*}[ht!]
    \renewcommand{\arraystretch}{1.7} 
    \centering
    \resizebox{\textwidth}{!}{
        \begin{tabular}{l ccc ccc ccc ccc}
            \toprule
            & \multicolumn{3}{c}{Close ($\phi = 0.9$, $\psi = 0.9$)} & \multicolumn{3}{c}{Medium ($\phi = 0.7$, $\psi = 0.7$)} & \multicolumn{3}{c}{Wide ($\phi = 0.5$, $\psi = 0.5$)} & \multicolumn{3}{c}{Very Wide ($\phi = 0.3$, $\psi = 0.3$)} \\  
            \cmidrule(lr){2-4}
            \cmidrule(lr){5-7}
            \cmidrule(lr){8-10}
            \cmidrule(lr){11-13}
            Method & PSNR $\uparrow$ & SSIM $\uparrow$ & LPIPS $\downarrow$ & PSNR $\uparrow$ & SSIM $\uparrow$ & LPIPS $\downarrow$ & PSNR $\uparrow$ & SSIM $\uparrow$ & LPIPS $\downarrow$ & PSNR $\uparrow$ & SSIM $\uparrow$ & LPIPS $\downarrow$  \\  
            \midrule
            \method (Ours) &
            \makecell{\textbf{19.66}\\(\textbf{14.72})} & \makecell{\textbf{0.757}\\-} & \makecell{\textbf{0.234}\\(\textbf{0.237})} & \makecell{\textbf{19.66}\\(\textbf{14.38})} & \makecell{\textbf{0.770}\\-} & \makecell{\textbf{0.229}\\(\textbf{0.243})} & \makecell{\textbf{19.41}\\(\textbf{13.72})} & \makecell{\textbf{0.783}\\-} & \makecell{\textbf{0.220}\\(\textbf{0.247})} & \makecell{\textbf{19.18}\\(\textbf{12.94})} & \makecell{\textbf{0.794}\\-} & \makecell{\textbf{0.209}\\(\textbf{0.258})} \\
            \midrule
            MASt3R (Point Cloud) & \makecell{18.56\\(13.57)} & \makecell{0.708\\-} & \makecell{0.278\\(0.283)} &  \makecell{18.51\\(12.96)} & \makecell{0.718\\-} & \makecell{0.259\\(0.280)} &  \makecell{18.73\\(12.50)} & \makecell{0.739\\-} & \makecell{0.245\\(0.293)} &  \makecell{18.44\\(11.27)} & \makecell{0.758\\-} & \makecell{0.242\\(0.322)} \\
            pixelSplat (MASt3R cams) & \makecell{15.48\\(10.53)} & \makecell{0.602\\-} & \makecell{0.439\\(0.447)} & \makecell{15.96\\(10.64)} & \makecell{0.648\\-} & \makecell{0.379\\(0.405)} & \makecell{15.94\\(10.14)} & \makecell{0.675\\-} & \makecell{0.343\\(0.394)} & \makecell{16.46\\(10.12)} & \makecell{0.708\\-} & \makecell{0.302\\(0.373)} \\
            pixelSplat (GT cams) & \makecell{15.67\\(10.71)} & \makecell{0.609\\-} & \makecell{0.436\\(0.443)} & \makecell{15.92\\(10.61)} & \makecell{0.643\\-} & \makecell{0.381\\(0.407)} & \makecell{16.08\\(10.33)} & \makecell{0.672\\-} & \makecell{0.407\\(0.392)} & \makecell{16.56\\(10.20)} & \makecell{0.709\\-} & \makecell{0.299\\(0.370)} \\
            \bottomrule 
        \end{tabular}
    }
    \caption{
        \textbf{Comparisons with the state of the art.}
        Performances are averaged over test scenes in ScanNet++.
        For each scene, the model takes two, unposed images as input and renders novel views for evaluation.
        \method shows improvements on all visual metrics.
    }
    \label{tab:scannet++}
\end{table*}
To address this, we introduce a loss masking strategy. For each target image, we calculate which pixels are visible in at least one of the context images. We unproject each point in the target image and reproject it onto each of the context images, checking if the rendered depth closely matches the ground truth depth. We show the construction of an example loss mask \cref{fig:loss_masks}.

Like existing generalized 3D-GS approaches~\cite{charatan2024pixelsplat, szymanowicz2024splatter, chen2024mvsplat}, we train using a \emph{weighted} combination of mean squared error loss (MSE) and perceptual similarity. Given, our rendered images ($\hat{\textbf{I}}$), ground truth images ($\textbf{I}$), and rendered loss masks $M$, the masked reconstruction loss is:
\begin{equation}
    \label{eqn:our_loss}
    \begin{split}
        L & = \lambda_{MSE} L_{MSE}(M \odot \hat{\textbf{I}}, M \odot \textbf{I}) \\
         & + \lambda_{LPIPS} L_{LPIPS}(M \odot \hat{\textbf{I}}, M \odot \textbf{I})
    \end{split}
\end{equation}
During training, existing methods~\cite{charatan2024pixelsplat,chen2024mvsplat,szymanowicz2024flash3d} assume that the images of each scene are in a video sequence. These methods use the number of frames between chosen context images as a proxy for the distance and overlap between the images, and select intermediary frames as the target frames for novel view synthesis supervision. We seek to generalize this approach to work with datasets of frames that are not in the form of a linear sequence, and to allow supervision from views that are not in-between the context images. During preprocessing, we calculate the overlap masks for each pair of images for each scene in the training set. During training, we select context images such that at least $\phi$\% of the pixels in the second image have direct correspondences in the first, and target images such that at least $\psi$\% of the pixels are present in at least one of the context images.
\section{Experimental Results}

Next, we describe our experimental setup (\cref{ssec:setup}), evaluate our method with a comparison to baselines (\cref{ssec:results}), and assess the significance of our model's components with an ablation study (\cref{ssec:ablation}).

\begin{figure*}
  \centering
  \includegraphics[width=1.0\textwidth]{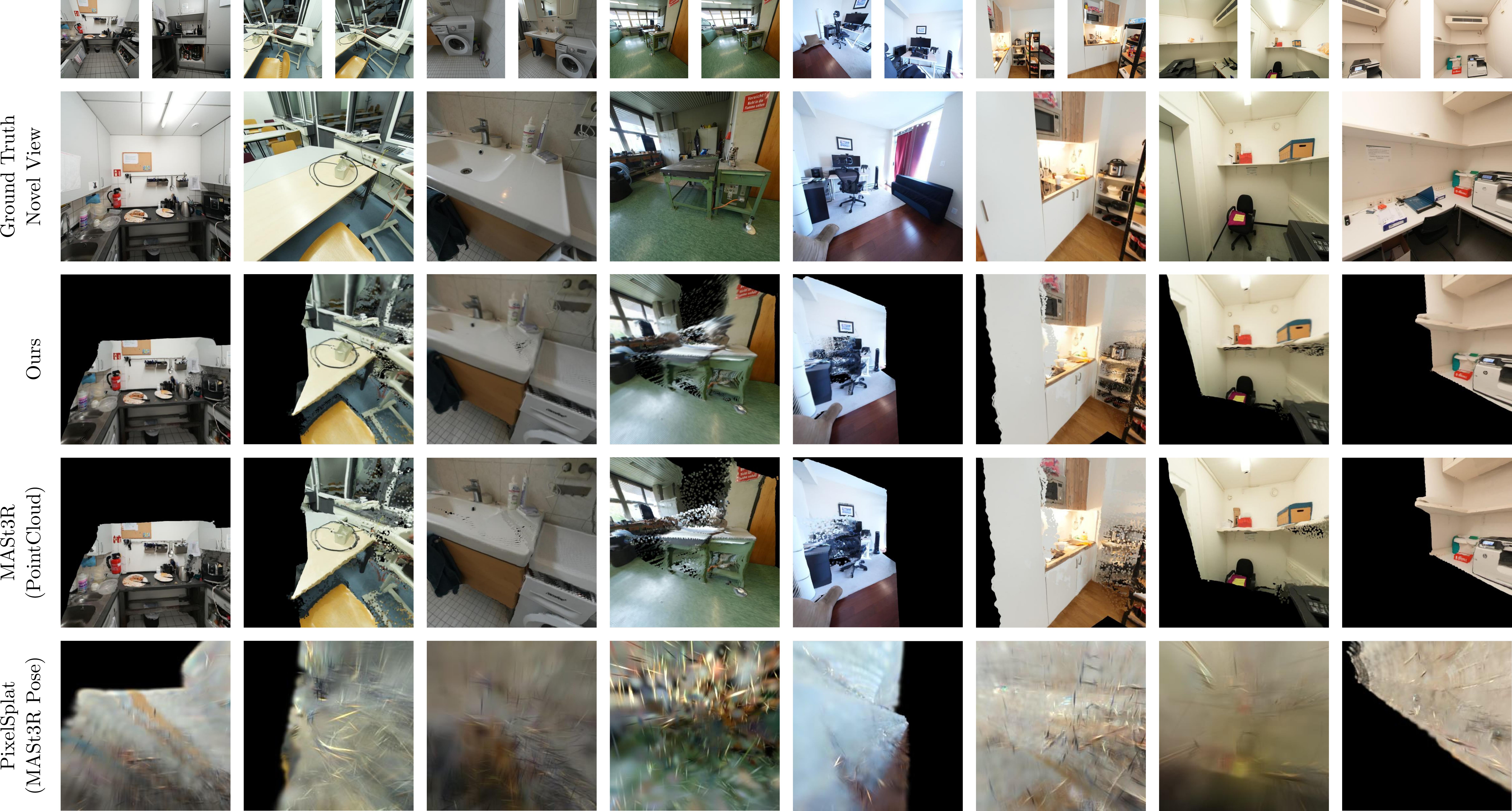}
  \caption{\textbf{Qualitative comparisons on ScanNet++.} We compare different methods on ScanNet++ testing examples. The two context camera views for each image are included in the first row of the table.
  } 
  \label{fig:comparison}
\end{figure*}
\begin{figure}[!ht]
  \centering
  \includegraphics[width=1.0\columnwidth]{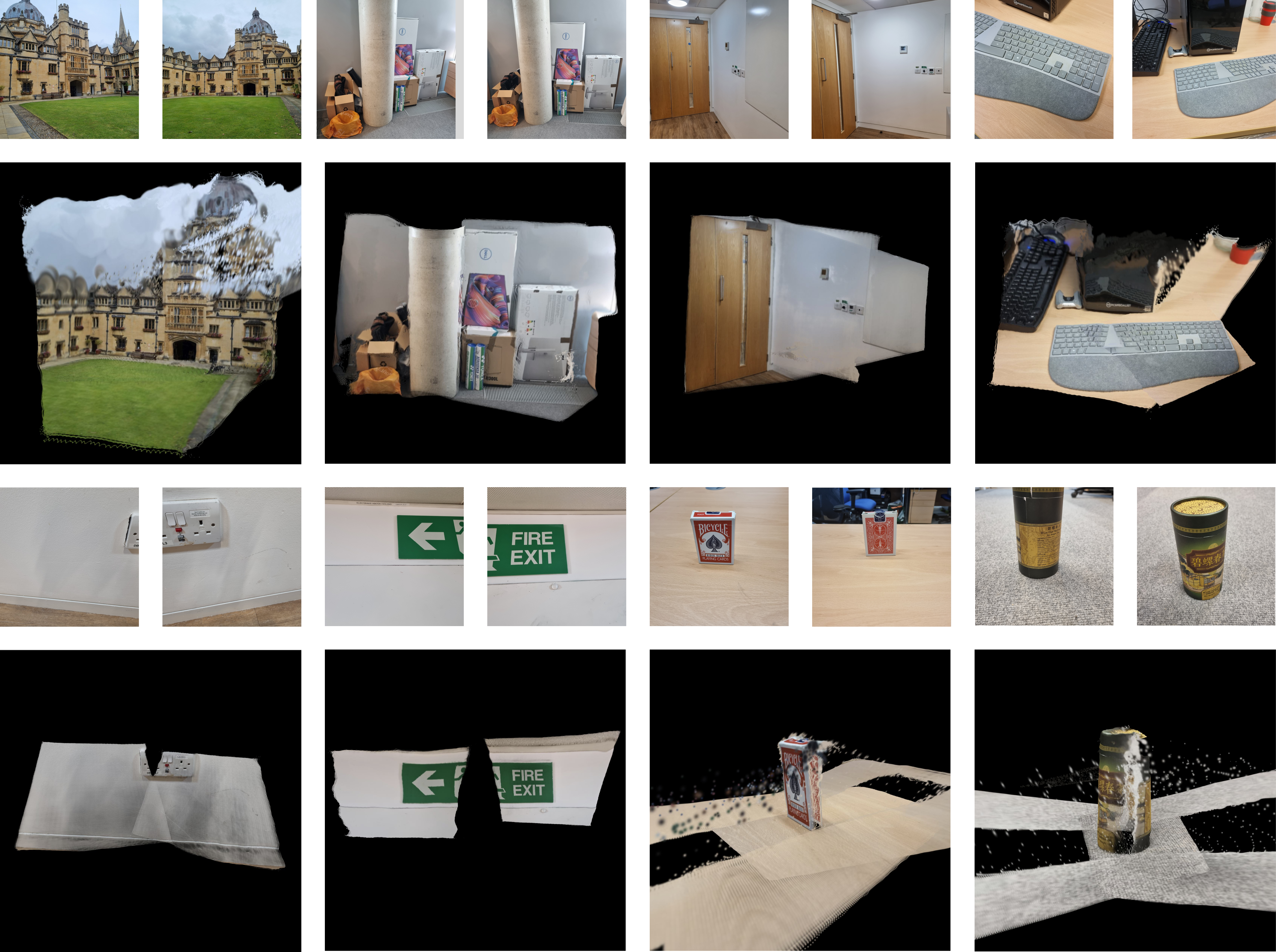}
  \caption{\textbf{Examples of Splatt3R generalizing to in-the-wild testing examples.} The bottom row showcases examples with few direct pixel correspondences between the two context images.
  } 
  \label{fig:comparison_in_the_wild}
\end{figure}

\subsection{Training and Evaluation Setup}
\label{ssec:setup}

\paragraph{Training details.} During each epoch, we randomly sample two input images, and three target images from each scene in the training split. As described in Section \ref{sec:protocols}, we select views using $\phi$ and $\psi$ parameters which we set at $\phi = \psi = 0.3$. We train our model for 2000 epochs ($\approx$ 500,000 iterations) at a resolution of $512 \times 512$, using $\lambda_{MSE} = 1.0$ and $\lambda_{LPIPS} = 0.25$. We optimize using the Adam optimizer at learning rate of $1.0 \times 10^{-5}$, with a weight decay of $0.05$, and a gradient clip value of $0.5$. 

\paragraph{Training data.}
We train our model using ScanNet++~\cite{yeshwanth2023scannet++}, which is a dataset of 450+ indoor scenes with ground truth depth obtained from high-resolution laser scans. We use the official ScanNet++ training and validation splits.

\paragraph{Testing datasets.}
We construct four testing subsets from ScanNet++ scenes to represent close together views (for high $\phi$ and $\psi$) and farther views with less overlap (for low $\phi$ and $\psi$).
The test scenes are not seen during training.
We ignore the frames marked as `bad' in the ScanNet++ dataset, and scenes that contain frames with no valid depth.

\paragraph{Metrics} are calculated after applying the loss masks to the rendered and target images. Metrics are reported both across the entire image, and averaging across just the pixels in the loss mask (in parenthesis for PSNR and LPIPS).

\paragraph{Baselines.}
To the best of our knowledge, \method is the first model that performs 3D reconstruction from a wide, unposed, stereo pair of images for novel view synthesis in a feed-forward manner. 
To evaluate our method, we construct baselines from existing works.
We test our method against directly rendering MASt3R's prediction as a colored point cloud, giving each point the color of its corresponding pixel. We wish to reconstruct and render the entire 3D scene, therefore we do not filter out points with low confidences from our point cloud renderings. We also compare our method against pixelSplat~\cite{charatan2024pixelsplat}, a generalizable 3D-GS reconstruction method that requires poses for reconstruction. We evaluate pixelSplat using ground truth camera poses, and also using camera poses estimated using the point clouds predicted from MASt3R. Please see the MASt3R paper for details on pose regression from MASt3R's predictions~\cite{leroy2024grounding}. We retrain baselines with the same dataloaders and training curricula where appropriate to present a fair comparison. Due to memory constraints when training pixelSplat, we train at 256x256, and initialize the model using the pretrained weights from the pixelSplat authors. We observe that when trained using the same data schedule, pixelSplat achieves very low accuracy. Therefore we adapt pixelSplat's curriculum learning strategy for our data, initially training the model at $\phi = \psi = 0.7$, and decreasing these values to $\phi = \psi = 0.3$ at the end of training.

\subsection{Results}
\label{ssec:results}

\paragraph{Quantitative evaluation.} We begin by reporting our quantitative results for ScanNet++ in \cref{tab:scannet++}. Our method outperforms both directly rendering the MASt3R point cloud, and reconstructing the scene using pixelSplat across all stereo baseline sizes. Critically, we find that our method outperforms pixelSplat even when pixelSplat is evaluated using the ground truth poses for each camera. When trained using the stereo baselines in our dataset, and when supervised from viewpoints which contain information not visible to the input cameras, we observe that the quality of reconstructions from pixelSplat significantly degrades.

\begin{table*}[ht!]
    \renewcommand{\arraystretch}{1.7} 
    \centering
    \resizebox{\textwidth}{!}{
        \begin{tabular}{l ccc ccc ccc ccc}
            \toprule
            & \multicolumn{3}{c}{Close ($\phi = 0.9$, $\psi = 0.9$)} & \multicolumn{3}{c}{Medium ($\phi = 0.7$, $\psi = 0.7$)} & \multicolumn{3}{c}{Wide ($\phi = 0.5$, $\psi = 0.5$)} & \multicolumn{3}{c}{Very Wide ($\phi = 0.3$, $\psi = 0.3$)} \\  
            \cmidrule(lr){2-4}
            \cmidrule(lr){5-7}
            \cmidrule(lr){8-10}
            \cmidrule(lr){11-13}
            Method & PSNR $\uparrow$ & SSIM $\uparrow$ & LPIPS $\downarrow$ & PSNR $\uparrow$ & SSIM $\uparrow$ & LPIPS $\downarrow$ & PSNR $\uparrow$ & SSIM $\uparrow$ & LPIPS $\downarrow$ & PSNR $\uparrow$ & SSIM $\uparrow$ & LPIPS $\downarrow$  \\  
            \midrule
            Ours &
            \makecell{19.66\\(14.72)} & \makecell{0.757\\-} & \makecell{0.234\\(0.237)} & \makecell{19.66\\(14.38)} & \makecell{0.770\\-} & \makecell{0.229\\(0.243)} & \makecell{19.41\\(13.72)} & \makecell{0.783\\-} & \makecell{0.220\\(0.247)} & \makecell{19.18\\(12.94)} & \makecell{0.794\\-} & \makecell{0.209\\(0.258)} \\
            \midrule
            + Finetune w/ MASt3R &
            \makecell{20.97\\(16.03)} & \makecell{0.780\\-} & \makecell{0.199\\(0.201)} & \makecell{20.41\\(15.13)} & \makecell{0.781\\-} & \makecell{0.214\\(0.226)} & \makecell{20.00\\(14.32)} & \makecell{0.793\\-} & \makecell{0.207\\(0.232)} & \makecell{19.69\\(13.45)} & \makecell{0.803\\-} & \makecell{0.197\\(0.241)} \\
            + Spherical Harmonics &
            \makecell{18.04\\(13.10)} & \makecell{0.730\\-} & \makecell{0.254\\(0.257)} & \makecell{18.57\\(13.29)} & \makecell{0.752\\-} & \makecell{0.248\\(0.259)} & \makecell{18.50\\(12.82)} & \makecell{0.768\\-} & \makecell{0.236\\(0.262)} & \makecell{18.40\\(12.16)} & \makecell{0.781\\-} & \makecell{0.226\\(0.272)} \\
            - LPIPS Loss &
            \makecell{19.62\\(14.68)} & \makecell{0.763\\-} & \makecell{0.277\\(0.282)} & \makecell{19.65\\(14.37)} & \makecell{0.776\\-} & \makecell{0.261\\(0.278)} & \makecell{19.41\\(13.73)} & \makecell{0.787\\-} & \makecell{0.245\\(0.278)} & \makecell{19.22\\(12.98)} & \makecell{0.797\\-} & \makecell{0.230\\(0.285)} \\
            - Offsets &
            \makecell{19.38\\(14.44)} & \makecell{0.757\\-} & \makecell{0.249\\(0.252)} & \makecell{19.25\\(13.97)} & \makecell{0.775\\-} & \makecell{0.242\\(0.256)} & \makecell{19.14\\(13.46)} & \makecell{0.792\\-} & \makecell{0.225\\(0.253)} & \makecell{19.09\\(12.85)} & \makecell{0.805\\-} & \makecell{0.209\\(0.255)} \\
            - Loss Masking &
            \makecell{N/A} & \makecell{N/A} & \makecell{N/A} & \makecell{N/A} & \makecell{N/A} & \makecell{N/A} & \makecell{N/A} & \makecell{N/A} & \makecell{N/A} & \makecell{N/A} & \makecell{N/A} & \makecell{N/A} \\
            \bottomrule 
        \end{tabular}
    }
    \caption{
        \textbf{Ablations on the ScanNet++ dataset.} When trained without loss masking, the memory requirements of rendering grow until training cannot continue.
    }
    \label{tab:ablation_table}
\end{table*}

\paragraph{Qualitative comparisons.} Next, we provide a qualitative comparison of each method using examples from ScanNet++ in \cref{fig:comparison}. We see that our method, like MASt3R is able to reconstruct the visible regions of the scene, while not attempting to reconstruct areas which are not visible to the context views. By masking our novel view rendering loss, our model does not learn to guess unseen regions of the scene. pixelSplat has a very poor reconstruction quality, visibly attempting to predict regions of the scene which cannot be seen from the input context views, and achieving poor accuracy even in reconstructable regions of the scene.
We also note the visual artifacts which are present when directly rendering the point clouds from MASt3R. Our learned 3D Gaussian representation is able to reduce the number of these artifacts, resulting in marginally higher quality renderings.

Here, we also note that our model is reconstructing the scene in metric scale. We can observe the accuracy of this scale prediction by noting how closely the viewpoint of the rendered image matches the ground truth image taken from that location. In only a few instances, such as the example in the third column, is there a significant misalignment between our rendered image and the ground truth image.

In \cref{fig:comparison_in_the_wild}, we attempt to generalize from our model, trained on ScanNet++, to real world data captured by a mobile phone. By only training our Gaussian head, we maintain MASt3R's ability to generalize to different scenes, such as the outdoor scene in the top left of the figure. Our predicted Gaussians are able to generalize from object-scale scenes up to large outdoor environments. We make a particular note of the bottom row of \cref{fig:comparison_in_the_wild}, where we show examples of reconstructing a scene from two images with little or no direct pixel correspondences, due to being taken directly side-by-side or from opposite sides of the same object. Traditional multi-view stereo systems based on image correspondences would fail in these scenarios, however MASt3R's data-driven approach allows these scenes to be reconstructed accurately.

\paragraph{Runtime comparisons.} 

\begin{table}[t!]
    \centering
    \renewcommand{\arraystretch}{1.2} 
    \scalebox{0.8}{
        \begin{tabular}{l cc}
            \toprule
            Method & Pose Est. & Encoding \\  
            \midrule
            Ours & - & 0.268 \\
            MASt3R (Point Cloud) & - & 0.263 \\
            PixelSplat (w/ MASt3R poses) & 10.72 & 0.156 \\
            \bottomrule 
        \end{tabular}
    }
    \caption{\textbf{Average time in seconds} required for position estimation (if relevant) and scene prediction.} 
    \label{tab:performance}
\end{table}

Next, we benchmark the time taken to reconstruct poses and perform scene reconstruction using each of these methods. Our method, and MASt3R, do not need to perform any explicit pose estimation, as all points and Gaussians are directly predicted in the same coordinate space. We see that our method can reconstruct scenes at  \raisebox{0.5ex}{\texttildelow}4 FPS on an RTX2080ti at 512x512 resolution. Because pixelSplat needs to use MASt3R and perform explicit point cloud alignment to estimate the poses of the images, our total runtime is significantly less than the time taken to estimate the poses for pixelSplat.

\subsection{Ablation studies}
\label{ssec:ablation}

In \cref{tab:ablation_table}, we run ablations on our method. We find that finetuning our MASt3R's 3D point predictions to ScanNet++ improves testing performance on ScanNet++, but we omit this finetuning from our other experiments for fair comparison with MASt3R. When training with spherical harmonics (with a degree of 4) instead of constant color Gaussians, we find that our performance decreases, likely due to overfitting spherical harmonics to our collection of training scenes. Like other works, we find that using an LPIPS loss term meaningfully increases the visual quality of the reconstructions. Our introduced offsets slightly improve performance across all metrics as well. Finally, if we omit our loss masking strategy, we find that the size of the Gaussians grows in an unbounded manner, until the memory cost of rendering the Gaussians causes training to halt.
\section{Conclusion}

We present \method, a feed-forward generalizable model for generating 3D Gaussian Splats from uncalibrated stereo images, without relying on camera intrinsics, extrinsics, or depth information. We find that simply using the MASt3R architecture to predict 3D Gaussian parameters, in combination with a loss-masking strategy during training, allows us to accurately reconstruct both 3D appearance and geometry from wide baselines. As we demonstrate in our experiments, \method outperforms both MASt3R and the current state-of-the-art in feed-forward splatting. 

{
    \small
    \bibliographystyle{ieeenat_fullname}
    \bibliography{main}

\begin{thebibliography}{69}
\providecommand{\natexlab}[1]{#1}
\providecommand{\url}[1]{\texttt{#1}}
\expandafter\ifx\csname urlstyle\endcsname\relax
  \providecommand{\doi}[1]{doi: #1}\else
  \providecommand{\doi}{doi: \begingroup \urlstyle{rm}\Url}\fi

\bibitem[Adelson and Bergen(1991)]{adelson1991plenoptic}
Edward~H Adelson and James~R Bergen.
\newblock \emph{The plenoptic function and the elements of early vision}.
\newblock MIT Press, 1991.

\bibitem[Barnard and Fischler(1982)]{barnard1982computational}
Stephen~T Barnard and Martin~A Fischler.
\newblock Computational stereo.
\newblock \emph{ACM Computing Surveys (CSUR)}, 1982.

\bibitem[Barron et~al.(2022)Barron, Mildenhall, Verbin, Srinivasan, and Hedman]{barron2022mipnerf360}
Jonathan~T Barron, Ben Mildenhall, Dor Verbin, Pratul~P Srinivasan, and Peter Hedman.
\newblock Mip-nerf 360: Unbounded anti-aliased neural radiance fields.
\newblock In \emph{CVPR}, 2022.

\bibitem[Bian et~al.(2023{\natexlab{a}})Bian, Bian, Prisacariu, and Torr]{bian2023porf}
Jia-Wang Bian, Wenjing Bian, Victor~Adrian Prisacariu, and Philip Torr.
\newblock Porf: Pose residual field for accurate neural surface reconstruction.
\newblock In \emph{ICLR}, 2023{\natexlab{a}}.

\bibitem[Bian et~al.(2023{\natexlab{b}})Bian, Wang, Li, Bian, and Prisacariu]{bian2023nope}
Wenjing Bian, Zirui Wang, Kejie Li, Jia-Wang Bian, and Victor~Adrian Prisacariu.
\newblock Nope-nerf: Optimising neural radiance field with no pose prior.
\newblock In \emph{CVPR}, 2023{\natexlab{b}}.

\bibitem[Chang and Chen(2018)]{chang2018pyramid}
Jia-Ren Chang and Yong-Sheng Chen.
\newblock Pyramid stereo matching network.
\newblock In \emph{CVPR}, 2018.

\bibitem[Charatan et~al.(2024)Charatan, Li, Tagliasacchi, and Sitzmann]{charatan2024pixelsplat}
David Charatan, Sizhe~Lester Li, Andrea Tagliasacchi, and Vincent Sitzmann.
\newblock pixelsplat: 3d gaussian splats from image pairs for scalable generalizable 3d reconstruction.
\newblock In \emph{CVPR}, 2024.

\bibitem[Chen et~al.(2021)Chen, Xu, Zhao, Zhang, Xiang, Yu, and Su]{chen2021mvsnerf}
Anpei Chen, Zexiang Xu, Fuqiang Zhao, Xiaoshuai Zhang, Fanbo Xiang, Jingyi Yu, and Hao Su.
\newblock Mvsnerf: Fast generalizable radiance field reconstruction from multi-view stereo.
\newblock In \emph{ICCV}, 2021.

\bibitem[Chen and Lee(2023)]{chen2023dbarf}
Yu Chen and Gim~Hee Lee.
\newblock Dbarf: Deep bundle-adjusting generalizable neural radiance fields.
\newblock In \emph{CVPR}, 2023.

\bibitem[Chen et~al.(2024)Chen, Xu, Zheng, Zhuang, Pollefeys, Geiger, Cham, and Cai]{chen2024mvsplat}
Yuedong Chen, Haofei Xu, Chuanxia Zheng, Bohan Zhuang, Marc Pollefeys, Andreas Geiger, Tat-Jen Cham, and Jianfei Cai.
\newblock Mvsplat: Efficient 3d gaussian splatting from sparse multi-view images.
\newblock \emph{ECCV}, 2024.

\bibitem[Chi et~al.(2021)Chi, Wang, Hao, Guo, and Yang]{chi2021feature}
Cheng Chi, Qingjie Wang, Tianyu Hao, Peng Guo, and Xin Yang.
\newblock Feature-level collaboration: Joint unsupervised learning of optical flow, stereo depth and camera motion.
\newblock In \emph{CVPR}, 2021.

\bibitem[Chibane et~al.(2021)Chibane, Bansal, Lazova, and Pons-Moll]{chibane2021stereo}
Julian Chibane, Aayush Bansal, Verica Lazova, and Gerard Pons-Moll.
\newblock Stereo radiance fields (srf): Learning view synthesis for sparse views of novel scenes.
\newblock In \emph{CVPR}, 2021.

\bibitem[Delaunoy and Pollefeys(2014)]{delaunoy2014photometric}
Ama{\"e}l Delaunoy and Marc Pollefeys.
\newblock Photometric bundle adjustment for dense multi-view 3d modeling.
\newblock In \emph{CVPR}, 2014.

\bibitem[DeTone et~al.(2018)DeTone, Malisiewicz, and Rabinovich]{detone2018superpoint}
Daniel DeTone, Tomasz Malisiewicz, and Andrew Rabinovich.
\newblock Superpoint: Self-supervised interest point detection and description.
\newblock In \emph{CVPRW}, 2018.

\bibitem[Dosovitskiy et~al.(2021)Dosovitskiy, Beyer, Kolesnikov, Weissenborn, Zhai, Unterthiner, Dehghani, Minderer, Heigold, Gelly, Uszkorei, and Houlsy]{dosovitskiy2020image}
Alexey Dosovitskiy, Lucas Beyer, Alexander Kolesnikov, Dirk Weissenborn, Xiaohua Zhai, Thomas Unterthiner, Mostafa Dehghani, Matthias Minderer, Georg Heigold, Sylvain Gelly, Jakob Uszkorei, and Neil Houlsy.
\newblock An image is worth 16x16 words: Transformers for image recognition at scale.
\newblock \emph{ICLR}, 2021.

\bibitem[Du et~al.(2023)Du, Smith, Tewari, and Sitzmann]{du2023cross}
Yilun Du, Cameron Smith, Ayush Tewari, and Vincent Sitzmann.
\newblock Learning to render novel views from wide-baseline stereo pairs.
\newblock In \emph{CVPR}, 2023.

\bibitem[Garg et~al.(2016)Garg, Bg, Carneiro, and Reid]{garg2016unsupervised}
Ravi Garg, Vijay~Kumar Bg, Gustavo Carneiro, and Ian Reid.
\newblock Unsupervised cnn for single view depth estimation: Geometry to the rescue.
\newblock In \emph{ECCV}, 2016.

\bibitem[Godard et~al.(2017)Godard, Mac~Aodha, and Brostow]{godard2017unsupervised}
Cl{\'e}ment Godard, Oisin Mac~Aodha, and Gabriel~J Brostow.
\newblock Unsupervised monocular depth estimation with left-right consistency.
\newblock In \emph{CVPR}, 2017.

\bibitem[Gortler et~al.(1996)Gortler, Grzeszczuk, Szeliski, and Cohen]{gortler1996lumigraph}
Steven~J Gortler, Radek Grzeszczuk, Richard Szeliski, and Michael~F Cohen.
\newblock The lumigraph.
\newblock In \emph{Computer Graphics and Interactive Techniques}, 1996.

\bibitem[Harris et~al.(1988)Harris, Stephens, et~al.]{harris1988combined}
Chris Harris, Mike Stephens, et~al.
\newblock A combined corner and edge detector.
\newblock In \emph{Alvey Vision Conference}, 1988.

\bibitem[Hartley and Schaffalitzky(2004)]{hartley2004sub}
Richard Hartley and Frederik Schaffalitzky.
\newblock L/sub/spl infin//minimization in geometric reconstruction problems.
\newblock In \emph{CVPR}, 2004.

\bibitem[Hartley and Sturm(1997)]{hartley1997triangulation}
Richard~I Hartley and Peter Sturm.
\newblock Triangulation.
\newblock \emph{Computer Vision and Image Understanding}, 1997.

\bibitem[Hartley et~al.(1992)Hartley, Gupta, and Chang]{hartley1992stereo}
Richard~I Hartley, Rajiv Gupta, and Tom Chang.
\newblock Stereo from uncalibrated cameras.
\newblock In \emph{CVPR}, 1992.

\bibitem[Ishikawa and Geiger(1998)]{ishikawa1998occlusions}
Hiroshi Ishikawa and Davi Geiger.
\newblock Occlusions, discontinuities, and epipolar lines in stereo.
\newblock In \emph{ECCV}, 1998.

\bibitem[Jeong et~al.(2021)Jeong, Ahn, Choy, Anandkumar, Cho, and Park]{jeong2021self}
Yoonwoo Jeong, Seokjun Ahn, Christopher Choy, Anima Anandkumar, Minsu Cho, and Jaesik Park.
\newblock Self-calibrating neural radiance fields.
\newblock In \emph{ICCV}, 2021.

\bibitem[Jiang et~al.(2023)Jiang, Jiang, Zhao, and Huang]{jiang2023leap}
Hanwen Jiang, Zhenyu Jiang, Yue Zhao, and Qixing Huang.
\newblock Leap: Liberate sparse-view 3d modeling from camera poses.
\newblock In \emph{ICLR}, 2023.

\bibitem[Johari et~al.(2022)Johari, Lepoittevin, and Fleuret]{johari2022geonerf}
Mohammad~Mahdi Johari, Yann Lepoittevin, and Fran{\c{c}}ois Fleuret.
\newblock Geonerf: Generalizing nerf with geometry priors.
\newblock In \emph{CVPR}, 2022.

\bibitem[Kanade et~al.(1996)Kanade, Yoshida, Oda, Kano, and Tanaka]{kanade1996stereo}
Takeo Kanade, Atsushi Yoshida, Kazuo Oda, Hiroshi Kano, and Masaya Tanaka.
\newblock A stereo machine for video-rate dense depth mapping and its new applications.
\newblock In \emph{CVPR}, 1996.

\bibitem[Kerbl et~al.(2023)Kerbl, Kopanas, Leimk{\"u}hler, and Drettakis]{kerbl2023gaussian}
Bernhard Kerbl, Georgios Kopanas, Thomas Leimk{\"u}hler, and George Drettakis.
\newblock 3d gaussian splatting for real-time radiance field rendering.
\newblock \emph{ToG}, 2023.

\bibitem[Lee et~al.(2024)Lee, Jin, Baek, and Cho]{lee2024generalizable}
Haechan Lee, Wonjoon Jin, Seung-Hwan Baek, and Sunghyun Cho.
\newblock Generalizable novel-view synthesis using a stereo camera.
\newblock \emph{CVPR}, 2024.

\bibitem[Leroy et~al.(2024)Leroy, Cabon, and Revaud]{leroy2024grounding}
Vincent Leroy, Yohann Cabon, and J{\'e}r{\^o}me Revaud.
\newblock Grounding image matching in 3d with mast3r.
\newblock \emph{arXiv preprint arXiv:2406.09756}, 2024.

\bibitem[Levoy and Hanrahan(1996)]{levoy1996light}
Marc Levoy and Pat Hanrahan.
\newblock Light field rendering.
\newblock In \emph{SIGGRAPH}, 1996.

\bibitem[Li et~al.(2024{\natexlab{a}})Li, Gao, Zhang, Wu, Dai, Zhao, Feng, Ding, Wang, and Han]{li2024ggrt}
Hao Li, Yuanyuan Gao, Dingwen Zhang, Chenming Wu, Yalun Dai, Chen Zhao, Haocheng Feng, Errui Ding, Jingdong Wang, and Junwei Han.
\newblock Ggrt: Towards generalizable 3d gaussians without pose priors in real-time.
\newblock \emph{ECCV}, 2024{\natexlab{a}}.

\bibitem[Li et~al.(2024{\natexlab{b}})Li, Gou, and Tan]{li2024taming}
Yaokun Li, Chao Gou, and Guang Tan.
\newblock Taming uncertainty in sparse-view generalizable nerf via indirect diffusion guidance.
\newblock \emph{arXiv preprint arXiv:2402.01217}, 2024{\natexlab{b}}.

\bibitem[Lin et~al.(2021)Lin, Ma, Torralba, and Lucey]{lin2021barf}
Chen-Hsuan Lin, Wei-Chiu Ma, Antonio Torralba, and Simon Lucey.
\newblock Barf: Bundle-adjusting neural radiance fields.
\newblock In \emph{ICCV}, 2021.

\bibitem[Liu et~al.(2024)Liu, Wang, Hu, Shen, Ye, Zang, Cao, Li, and Liu]{liu2024fast}
Tianqi Liu, Guangcong Wang, Shoukang Hu, Liao Shen, Xinyi Ye, Yuhang Zang, Zhiguo Cao, Wei Li, and Ziwei Liu.
\newblock Fast generalizable gaussian splatting reconstruction from multi-view stereo.
\newblock \emph{ECCV}, 2024.

\bibitem[Liu et~al.(2022)Liu, Peng, Liu, Wang, Wang, Theobalt, Zhou, and Wang]{liu2022neural}
Yuan Liu, Sida Peng, Lingjie Liu, Qianqian Wang, Peng Wang, Christian Theobalt, Xiaowei Zhou, and Wenping Wang.
\newblock Neural rays for occlusion-aware image-based rendering.
\newblock In \emph{CVPR}, 2022.

\bibitem[Long et~al.(2022)Long, Lin, Wang, Komura, and Wang]{long2022sparseneus}
Xiaoxiao Long, Cheng Lin, Peng Wang, Taku Komura, and Wenping Wang.
\newblock Sparseneus: Fast generalizable neural surface reconstruction from sparse views.
\newblock In \emph{ECCV}, 2022.

\bibitem[Lowe(2004)]{lowe2004distinctive}
David~G Lowe.
\newblock Distinctive image features from scale-invariant keypoints.
\newblock \emph{IJCV}, 2004.

\bibitem[Luong and Faugeras(1996)]{luong1996fundamental}
Quan-Tuan Luong and Olivier~D Faugeras.
\newblock The fundamental matrix: Theory, algorithms, and stability analysis.
\newblock \emph{IJCV}, 1996.

\bibitem[Mildenhall et~al.(2020)Mildenhall, Srinivasan, Tancik, Barron, Ramamoorthi, and Ng]{mildenhall2020nerf}
Ben Mildenhall, Pratul~P. Srinivasan, Matthew Tancik, Jonathan~T. Barron, Ravi Ramamoorthi, and Ren Ng.
\newblock Nerf: Representing scenes as neural radiance fields for view synthesis.
\newblock In \emph{ECCV}, 2020.

\bibitem[M{\"u}ller et~al.(2022)M{\"u}ller, Evans, Schied, and Keller]{muller2022instant}
Thomas M{\"u}ller, Alex Evans, Christoph Schied, and Alexander Keller.
\newblock Instant neural graphics primitives with a multiresolution hash encoding.
\newblock In \emph{SIGGRAPH}, 2022.

\bibitem[Ni et~al.(2024)Ni, Yang, Yang, Wang, Ma, and Kwong]{ni2024colnerf}
Zhangkai Ni, Peiqi Yang, Wenhan Yang, Hanli Wang, Lin Ma, and Sam Kwong.
\newblock Colnerf: Collaboration for generalizable sparse input neural radiance field.
\newblock In \emph{AAAI}, 2024.

\bibitem[Ranftl and Koltun(2018)]{ranftl2018deep}
Ren{\'e} Ranftl and Vladlen Koltun.
\newblock Deep fundamental matrix estimation.
\newblock In \emph{ECCV}, 2018.

\bibitem[Ranftl et~al.(2021)Ranftl, Bochkovskiy, and Koltun]{ranftl2021vision}
Ren{\'e} Ranftl, Alexey Bochkovskiy, and Vladlen Koltun.
\newblock Vision transformers for dense prediction.
\newblock In \emph{ICCV}, 2021.

\bibitem[Reizenstein et~al.(2021)Reizenstein, Shapovalov, Henzler, Sbordone, Labatut, and Novotny]{reizenstein2021common}
Jeremy Reizenstein, Roman Shapovalov, Philipp Henzler, Luca Sbordone, Patrick Labatut, and David Novotny.
\newblock Common objects in 3d: Large-scale learning and evaluation of real-life 3d category reconstruction.
\newblock In \emph{ICCV}, 2021.

\bibitem[Sitzmann et~al.(2019)Sitzmann, Zollh{\"o}fer, and Wetzstein]{sitzmann2019scene}
Vincent Sitzmann, Michael Zollh{\"o}fer, and Gordon Wetzstein.
\newblock Scene representation networks: Continuous 3d-structure-aware neural scene representations.
\newblock \emph{NeurIPS}, 2019.

\bibitem[Sitzmann et~al.(2021)Sitzmann, Rezchikov, Freeman, Tenenbaum, and Durand]{sitzmann2021light}
Vincent Sitzmann, Semon Rezchikov, Bill Freeman, Josh Tenenbaum, and Fredo Durand.
\newblock Light field networks: Neural scene representations with single-evaluation rendering.
\newblock \emph{NeurIPS}, 2021.

\bibitem[Smith et~al.(2023)Smith, Du, Tewari, and Sitzmann]{smith2023flowcam}
Cameron Smith, Yilun Du, Ayush Tewari, and Vincent Sitzmann.
\newblock Flowcam: Training generalizable 3d radiance fields without camera poses via pixel-aligned scene flow.
\newblock In \emph{NeurIPS}, 2023.

\bibitem[Suhail et~al.(2022)Suhail, Esteves, Sigal, and Makadia]{suhail2022generalizable}
Mohammed Suhail, Carlos Esteves, Leonid Sigal, and Ameesh Makadia.
\newblock Generalizable patch-based neural rendering.
\newblock In \emph{ECCV}, 2022.

\bibitem[Szymanowicz et~al.(2024{\natexlab{a}})Szymanowicz, Insafutdinov, Zheng, Campbell, Henriques, Rupprecht, and Vedaldi]{szymanowicz2024flash3d}
Stanislaw Szymanowicz, Eldar Insafutdinov, Chuanxia Zheng, Dylan Campbell, Jo{\~a}o~F Henriques, Christian Rupprecht, and Andrea Vedaldi.
\newblock Flash3d: Feed-forward generalisable 3d scene reconstruction from a single image.
\newblock \emph{arXiv preprint arXiv:2406.04343}, 2024{\natexlab{a}}.

\bibitem[Szymanowicz et~al.(2024{\natexlab{b}})Szymanowicz, Rupprecht, and Vedaldi]{szymanowicz2024splatter}
Stanislaw Szymanowicz, Chrisitian Rupprecht, and Andrea Vedaldi.
\newblock Splatter image: Ultra-fast single-view 3d reconstruction.
\newblock In \emph{CVPR}, 2024{\natexlab{b}}.

\bibitem[Trajkovi{\'c} and Hedley(1998)]{trajkovic1998fast}
Miroslav Trajkovi{\'c} and Mark Hedley.
\newblock Fast corner detection.
\newblock \emph{Image and Vision Computing}, 1998.

\bibitem[Truong et~al.(2023)Truong, Rakotosaona, Manhardt, and Tombari]{truong2023sparf}
Prune Truong, Marie-Julie Rakotosaona, Fabian Manhardt, and Federico Tombari.
\newblock Sparf: Neural radiance fields from sparse and noisy poses.
\newblock In \emph{CVPR}, 2023.

\bibitem[Wang et~al.(2021{\natexlab{a}})Wang, Zhong, Dai, Birchfield, Zhang, Smolyanskiy, and Li]{wang2021deep}
Jianyuan Wang, Yiran Zhong, Yuchao Dai, Stan Birchfield, Kaihao Zhang, Nikolai Smolyanskiy, and Hongdong Li.
\newblock Deep two-view structure-from-motion revisited.
\newblock In \emph{CVPR}, 2021{\natexlab{a}}.

\bibitem[Wang et~al.(2021{\natexlab{b}})Wang, Wang, Genova, Srinivasan, Zhou, Barron, Martin-Brualla, Snavely, and Funkhouser]{wang2021ibrnet}
Qianqian Wang, Zhicheng Wang, Kyle Genova, Pratul Srinivasan, Howard Zhou, Jonathan~T. Barron, Ricardo Martin-Brualla, Noah Snavely, and Thomas Funkhouser.
\newblock Ibrnet: Learning multi-view image-based rendering.
\newblock In \emph{CVPR}, 2021{\natexlab{b}}.

\bibitem[Wang et~al.(2024)Wang, Leroy, Cabon, Chidlovskii, and Revaud]{wang2024dust3r}
Shuzhe Wang, Vincent Leroy, Yohan Cabon, Boris Chidlovskii, and Jerome Revaud.
\newblock Dust3r: Geometric 3d vision made easy.
\newblock In \emph{CVPR}, 2024.

\bibitem[Wang et~al.(2021{\natexlab{c}})Wang, Wu, Xie, Chen, and Prisacariu]{wang2021nerfmm}
Zirui Wang, Shangzhe Wu, Weidi Xie, Min Chen, and Victor~Adrian Prisacariu.
\newblock Nerf--: Neural radiance fields without known camera parameters.
\newblock \emph{arXiv preprint arXiv:2102.07064}, 2021{\natexlab{c}}.

\bibitem[Wewer et~al.(2024)Wewer, Raj, Ilg, Schiele, and Lenssen]{wewer2024latentsplat}
Christopher Wewer, Kevin Raj, Eddy Ilg, Bernt Schiele, and Jan~Eric Lenssen.
\newblock latentsplat: Autoencoding variational gaussians for fast generalizable 3d reconstruction.
\newblock \emph{ECCV}, 2024.

\bibitem[Woodford and Rosten(2020)]{woodford2020large}
Oliver~J Woodford and Edward Rosten.
\newblock Large scale photometric bundle adjustment.
\newblock In \emph{BMVC}, 2020.

\bibitem[Yeshwanth et~al.(2023)Yeshwanth, Liu, Nie{\ss}ner, and Dai]{yeshwanth2023scannet++}
Chandan Yeshwanth, Yueh-Cheng Liu, Matthias Nie{\ss}ner, and Angela Dai.
\newblock Scannet++: A high-fidelity dataset of 3d indoor scenes.
\newblock In \emph{ICCV}, 2023.

\bibitem[Yin and Shi(2018)]{yin2018geonet}
Zhichao Yin and Jianping Shi.
\newblock Geonet: Unsupervised learning of dense depth, optical flow and camera pose.
\newblock In \emph{CVPR}, 2018.

\bibitem[Yu et~al.(2021)Yu, Ye, Tancik, and Kanazawa]{yu2021pixelnerf}
Alex Yu, Vickie Ye, Matthew Tancik, and Angjoo Kanazawa.
\newblock pixelnerf: Neural radiance fields from one or few images.
\newblock In \emph{CVPR}, 2021.

\bibitem[Zbontar and LeCun(2015)]{zbontar2015computing}
Jure Zbontar and Yann LeCun.
\newblock Computing the stereo matching cost with a convolutional neural network.
\newblock In \emph{CVPR}, 2015.

\bibitem[Zhan et~al.(2018)Zhan, Garg, Weerasekera, Li, Agarwal, and Reid]{zhan2018unsupervised}
Huangying Zhan, Ravi Garg, Chamara~Saroj Weerasekera, Kejie Li, Harsh Agarwal, and Ian Reid.
\newblock Unsupervised learning of monocular depth estimation and visual odometry with deep feature reconstruction.
\newblock In \emph{CVPR}, 2018.

\bibitem[Zhang et~al.(2019)Zhang, Prisacariu, Yang, and Torr]{zhang2019ga}
Feihu Zhang, Victor Prisacariu, Ruigang Yang, and Philip~HS Torr.
\newblock Ga-net: Guided aggregation net for end-to-end stereo matching.
\newblock In \emph{CVPR}, 2019.

\bibitem[Zhang et~al.(1995)Zhang, Deriche, Faugeras, and Luong]{zhang1995robust}
Zhengyou Zhang, Rachid Deriche, Olivier Faugeras, and Quang-Tuan Luong.
\newblock A robust technique for matching two uncalibrated images through the recovery of the unknown epipolar geometry.
\newblock \emph{Artificial Intelligence}, 1995.

\bibitem[Zheng et~al.(2024)Zheng, Zhou, Shao, Liu, Zhang, Nie, and Liu]{zheng2024gpsgaussian}
Shunyuan Zheng, Boyao Zhou, Ruizhi Shao, Boning Liu, Shengping Zhang, Liqiang Nie, and Yebin Liu.
\newblock Gps-gaussian: Generalizable pixel-wise 3d gaussian splatting for real-time human novel view synthesis.
\newblock In \emph{CVPR}, 2024.

\bibitem[Zhou et~al.(2018)Zhou, Tucker, Flynn, Fyffe, and Snavely]{zhou2018stereo}
Tinghui Zhou, Richard Tucker, John Flynn, Graham Fyffe, and Noah Snavely.
\newblock Stereo magnification: Learning view synthesis using multiplane images.
\newblock In \emph{SIGGRAPH}, 2018.

\end{thebibliography}
}

\end{document}